\documentclass{article}
\usepackage{PRIMEarxiv}
\usepackage[utf8]{inputenc} 
\usepackage[T1]{fontenc}    
\usepackage{hyperref}       
\usepackage{url}            
\usepackage{booktabs}       
\usepackage{amsfonts}       
\usepackage{nicefrac}       
\usepackage{microtype}      
\usepackage{lipsum}
\usepackage{fancyhdr}       
\usepackage{graphicx}  
\usepackage{arabtex}
\graphicspath{{media/}}     

\title{Noor-Ghateh: A Benchmark Dataset
for Evaluating Arabic Word Segmentation Tools
in Hadith Domain}

\author{
Huda AlShuhayeb, Behrouz Minaei–Bidgoli \\
  Iran University of Science and Technology \\
  \texttt{hudaalshuhayeb@gmail.com,
   b\_minaei@iust.ac.ir} \\
   \And
  Mohammad E. Shenassa,Sayyed-Ali Hossayni \\
  Computer Research Center of Islamic Sciences (CRCIS) \\
\texttt{eshenassa@gmail.com,
hossayni@iran.ir} \\
}

\begin{document}
\maketitle

\section*{Abstract}
There are numerous complex and rich morphological features in the Arabic language, which are highly useful when analyzing traditional Arabic textbooks, especially in the literary and religious contexts, and help in understanding the meaning of the textbooks.
Vocabulary separation means separating the word into different components, such as the root and affixes. In the morphological datasets, the variety of markers and the number of data samples help to evaluate the morphological techniques.
In this paper, we present a standard dataset for analyzing the Arabic segmentation tools, which includes approximately 223,690 words from the "Shariat al–Islam" book, labeled by human experts. In terms of volume and word variety, this dataset is superior to the other Hadith Arabic datasets, to the best of our knowledge.
To estimate the dataset, we applied different methods, including Farasa, Camel, and ALP and reported the annotation quality, and analyzed the benchmark specifications, as well. This benchmark is considered to become the primary benchmark of Hadith segmentation task, from now on.


\section{Introduction}
Recently, the Arabic language has gained popularity among researchers because it is the main language in many countries, and the implementation of various Arabic systems can meet user needs. Arabic has complex semantic and phonetic structures,
as well as a rich morphology, which makes the analysis process challenging. 
Morphological analysis is incredibly important in NLP tasks such as search, spelling check, and linguistics, and it serves as the first step in any syntactic analysis. 
The structure of Arabic words is based on roots which create derivative words by adding prefixes or suffixes to form verbs, adjectives, and nouns. 
\par
The derivation of a word from a root typically occurs in two steps. 
First, a stem is produced by applying a template to the root. 
Then, in the second step, affixes are added to the stem.
In addition, stemming algorithms can also significantly reduce the grammatical structure of an inflected or derived word. 
In English, affixes are ordinarily divided into prefixes and suffixes. 
In Arabic, there are three types of affixes: prefixes, infixes, and suffixes.
\par
As a result, existing methods cannot always extract stems or roots of words with infixes.
Data scarcity is a significant challenge in Arabic NLP, due to the language's morphological richness and lack of standardized orthographic rules.
Other challenges include resolving ambiguities, such as anomalies in roots (most of which are composed of three letters), orthographic variations, complex morphology, and highly rich.
Morphology is essential for many NLP tasks, making morphological analysis the starting point for most language processing systems aimed at facilitating human-computer interaction.
\par
In morphological analyzers, all possible analyses are provided out of context for a given word.
To achieve optimal results, morphological analyzers must return all possible segments of a given word and account for the various inflected forms of its lemma.
To strengthen these analyzers, existing linguistic dictionaries are used and manual checks are performed.
It is difficult to find annotated Arabic corpora that are widely accessible. 
\par
In this paper, with the development of the Noor-Ghateh Dataset, we aim to enhance Arabic corpus resources.
Accordingly, we have begun meticulously annotating the book Shariat al-Islam, which is a collection of Hadith.
Research in Arabic linguistic processing requires annotated linguistic resources, such as a Hadith corpus.
\par
To the best of our knowledge, there are currently no morphological datasets specific to Arabic Hadiths available.
In addition to enriching Arabic corpus resources, the Noor-Ghateh corpora can be integrated with various Arabic language processing tools.
\par
This paper is structured as follows.
The related work section is dedicated to reviewing the corresponding literature.
Then, a separate section is dedicated to introducing the datasets, including the previously published ones, the presented one (Noor-Ghateh), and their comparison.
Thereafter, some stemming algorithm experiments are proposed to provide a comparison between the manner of different algorithms in this dataset, alongside the other existing ones.
Finally, the conclusion section finished the paper.

\section{Related Works}
Over the past few decades, several methods have been proposed for Arabic morphological analysis.
Morphological analysis engines still face many challenges.
This section discusses recent research on Arabic morphological analysis.
\par
One of the early morphological analysis engines was Buckwalter \cite{buckwalter2002buckwalter}, which was designed for Arabic text POS tagging.
The data consisted essentially of three Arabic–English lexicon files. Researchers who require Arabic lexicons and morphological analysis can use this resource. 
The designers created six tables: three lexical tables and three compatibility tables. 
The system supported root-pattern morphology and the first version was produced by LDC.
BAMA Version 2.0 \cite{buckwalter2004buckwalter} was released in 2004. 
Its latest version of this system (called SAMA) \cite{graff2010standard} corrected several inconsistencies and added new features while revising entries to match updated standards, but it also introduced new inconsistencies.
The LDC released the PATB \cite{maamouri2004penn}, which supported numerous new morphological modeling studies in Arabic. 
The PATB relies on the existence of BAMA and SAMA. 
\par
Also, the paper \cite{elghamry2004constraint} identifies Arabic roots using supervised learning methods that rely only on rule restrictions. 
This algorithm was tested on a raw Arabic text collection extracted from the Aljazeera website. They tested their work on 2,700 unique words and achieved an accuracy of 92 \%.
This algorithm was tested on an unchanged Arabic textbook collection uprooted from the Aljazeera website and they tested their work on 2,700 unique words and reached an accuracy of 92 \%. 
An automated feature covering rationality was implemented, though it was not fully examined.
\par
Another morphological analysis tool is MAGEAD \cite{habash2006magead}, which focuses on modeling both Dialectal Arabic (DA) and Modern Standard Arabic (MSA). 
It can analyze the structure of colloquial words and MSA words based on their root-pattern associations and morph, or generate forms in reverse.
It supports root-pattern-based online parsing, separates phonological and orthographic representations, and combines terms from different dialects.
It includes roots with varying degrees of accuracy. 
The rationality feature was automatically populated, but it was not thoroughly examined. 
\par
The paper \cite{rodrigues2007learning} uses only statistical methods, semitic root morphology, and no lexicon to develop an Arabic morphological approach.
The authors introduced a negative evidence-based Arabic concatenative morphology algorithm within a root identification process that removes certain vowel patterns as potential root indicators. 
The ALMORGEANA \cite{habash2007arabic} morphological analysis tool pre-compiles the various components required by the system. 
In its database, feature definitions are specified along with default characteristics for each POS label, but phonological forms are not covered.
Although it covers certain rationality features, these were acquired automatically and were not thoroughly examined. 
\par
ElixirFM \cite{smrz2007elixirfm} reused the Buckwalter lexicon and supported functional gender, number, full case, and state modeling, as well as phonological forms. 
The authors developed a domain-specific language embedded in Haskell, containing verbal definitions that simultaneously act as source code.
They also prepared a data structure, ready for independent processing and editing. 
\par
In the paper \cite{daya2008identifying}, the authors applied several machine learning methods to \cite{roth1998learning}, a multi-marker classifier specifically designed for learning in various disciplines. 
The model \cite{snyder2008unsupervised} extracts cross-lingual patterns in unsupervised morphological segmentation. 
The authors also provide evidence that considering closely related languages can be more effective than unrelated language pairs. 
The paper \cite{poon2009unsupervised} introduced a logarithmic-linear model for unsupervised morphological segmentation, as well as learning and inference algorithms involving counter-slice estimation. 
The system supports only monolingual specifications.
\par
An unvoweled Arabic text morphological analysis system is presented in the paper \cite{boudlal2011markovian}. 
As a first step, the authors tested whether the primary module was effective in identifying roots assigned by evaluators and observed that the system generated between one to twelve roots.
As the first step, they tested whether the primary module was effective in searching out roots assigned by evaluators, and observed that from one to twelve roots were generated by the system.
Additionally, by introducing an approach supported by Hidden Markov Models, they identified the correct root for each word.
\par
The paper \cite{attia2011open} used machine literacy to prognosticate morphosyntactic features. 
For broken plurals, the authors used Levenshtein Distance and Arabic word pattern matching. 
The approach of \cite{fullwood2013learning} is based on Bayesian language modeling, presenting a model of loanword lexicon learning capable of handling continuous and unrelated morphological changes up to the level of two coinages.

The paper \cite{khaliq2013induction} presented an unsupervised approach to learning morphology.
It focuses on learning tri-partite roots and pattern templates.
CALIMAGLF \cite{khalifa2017morphological} is a morphological analyzer for Gulf Arabic, created from entries in a phonetic lexicon, along with orthographic paradigms and their associated variants.
Another system for Gulf Arabic \cite{khalifa2020morphological} also utilized morphological analyzers, disambiguation models, and different sizes of training data. 
\par
Also, the Annotated Gumar Corpus \cite{taji2018arabic} is proposed, as a subset of the Gumar corpus in Emirati Arabic \cite{khalifa2016large}, and three morphological analyzers are employed.
They include SAMA \cite{graff2010standard}, CALIMA \cite{habash2012morphological}, and GLF-MAPC for Gulf Arabic, all automatically generated.
In \cite{taji2018arabic}, the Annotated Gumar Corpus is used, as a subset of the Gumar corpus in Emirati Arabic \cite{khalifa2016large} and three morphological analyzers (SAMA \cite{graff2010standard} CALIMA \cite{habash2012morphological} and GLF-MAPC) are used which are automatically generated by sample achievement. 
\par
The paper \cite{gridach2011developing} developed a morphological analysis and generation system for Modern Standard Arabic.
Their system relies on Arabic morphological automaton technology that identifies roots and patterns to create a set of morphological automata, which can also be immediately used for Arabic morphological analysis and generation. 

The paper \cite{zalmout2017don} is based on Recurrent Neural Networks (RNNs). 
The authors prepare Long Short-Term Memory (LSTM) blocks in varied forms and embedding sets to model different morphological characteristics.

Alkhalil \cite{boudlal2010alkhalil}\cite{boudchiche2017alkhalil} is a transformative syntactic analyzer for standard Arabic words with two versions.
This system is based on stem-based morphology, including root-pattern and syntactic features, and can process unvoweled texts.
It supports the modeling of a comprehensive set of Arabic morphological rules.
The system used the Tashkeela and Nemlar corpora and worked with non-vowelized, partially vowelized, or fully vowelized texts. 
\par
In the paper \cite{Zribi2013Morphological}, the derivative patterns are modified for Tunisian, and Tunisian-specific roots and patterns are added.
The authors of the paper \cite{alkuhlani2011corpus} used a part of the PATB to incorporate functional gender, numbers, and grammatical agreement, but not the entire BAMA or SAMA database. 
\par
MADAMIRA \cite{pasha2014madamira} is another system that combines MADA \cite{habash2005arabic},\cite{habash2009mada+} and AMIRA \cite{diab2007automatic} for morphological analysis and disambiguation of Arabic words. 
Initially, the system analyzes words in the sentence out of context using the SAMA analyzer, then uses SVMs to disambiguate between the words obtained from the previous step.
MADAMIRA handles disambiguation and tokenization differently.
\par
Basma \cite{alansary2016basma} used BAMA for the morphological disambiguation process. 
The authors modified the number, gender, and definiteness to align with their morphosyntactic properties. 
They also added some markers to the ICA lexicon, as well as additional lemmas and glossaries. 
A new analysis and qualifier were also added, including root, stem pattern, and named entity analysis.
It also determines the lemmas, roots, stem patterns, figures, genders, definiteness, case, and eventually the communication of all words, grounded on the most effective POS result. 
Modified in BAMA’s AraMorph Perl script, which was compatible with the recently added features in BAMA’s output, root, and pattern. 
They separated the prefixes and suffixes from the stem, displayed each input word, and showed the output of BAMA for words that did not have any results. 
\par
YAMAMA \cite{khalifa2016yamama} worked like MADAMIRA in analysis and disambiguation. 
For disambiguation, a maximum likelihood model is used along with a word lookup model for morphological analysis.
They used the Penn Arabic Treebank (PATB sections 1, 2, and 3) and, for EGY, the ARZ Treebank. 
CALIMA \cite{habash2012morphological} extends the ECAL \cite{kilany2002egyptian} and designed a morphological analyzer for Egyptian lanuage that presents a linguistically accurate, large-scale model. 
Multiple orthographic variants are accepted and regularized to traditional orthography.
There are 100K stems and 36K lemmas in CALIMA, including 1,179 suffixes and 2,421 prefixes. 
The total number of words analyzable by CALIMA is 48 million words. 
The CALIMA–Star \cite{taji2018arabic} system supports functional and form-based morphological characteristics, including built-in tokenization, phonological representation, and verbal rationality.
\par
The paper \cite{alshargi2019morphologically} presented a set of morphological reflections that provided an aggregate of more than 200,000 words for seven Arabic dialects, which were manually annotated according to common norms for spelling, syntactic structures, punctuation, morphological units, and English words.
\par
The paper \cite{moulay2023development} addresses the unique challenges posed by the Algerian dialect, such as code-switching, agglutination, and syntactic flexibility. 
The authors developed a tool called TOKAD for text segmentation, using both rule-based and machine-learning approaches. 
The authors also created a segmented corpus named DzTOK to support their work. 
\par
Also, SALMA \cite{sawalha2013salma} employs a combination of empirical and rule-based methods for morphological analysis. The authors conducted an empirical study to apply standards and tools for Arabic morphological analysis. 
This involved using linguistic information extracted from traditional Arabic grammar books.
The authors developed a set of rules to capture the morphological features of Arabic.
These rules were designed to handle both vowelized and non-vowelized text, ensuring comprehensive coverage of Arabic morphology.
\par
The paper \cite{himdi2023tasaheel} introduces Tasaheel, a comprehensive tool designed for Arabic textual analysis. 
It integrates various NLP tasks, including stemming, segmentation, normalization, NER, and part-of-speech tagging. Tasaheel offers two main utilities: Traditional NLP Tasks and Advanced Textual Analysis.
Traditional NLP Tasks include fundamental tasks like stemming, segmentation, normalization, NER, and part-of-speech tagging, utilizing open-source Arabic NLP packages. 
Advanced Textual Analysis includes unique functionalities such as tag summaries for parts of speech, emotion, polarity, linguistics, and domain-specific words. It also features affix extraction, data management, and search capabilities.
\par
The paper \cite{nazih2024ibn} introduces a new morphological analyzer designed to enhance the processing of Arabic language.
This tool builds on the strengths of the Buckwalter BAMA, aiming to improve both speed and quality. 
The Ibn-Ginni morphological analyzer offers several key improvements over the Buckwalter BAMA. It can analyze approximately 0.6 million more words than BAMA, significantly enhancing its coverage of the Arabic language, and is faster, with an average word analysis time of 0.3 milliseconds. 
It inherits the speed and quality of BAMA but addresses its limitations, particularly in covering classical Arabic.
\par
The paper \cite{qarah2024comprehensive} evaluates the performance of different tokenizers, including WordPiece, SentencePiece, and Byte-Pair Encoding (BBPE), on Arabic large language models (LLMs). 
The authors pretrained three BERT models using each tokenizer and measured their performance across seven NLP tasks using 29 datasets.
They found that the model using the SentencePiece tokenizer significantly outperformed the others.

\section{Noor-Ghateh DataSet}
In this section, we first review the existing Arabic datasets.
Then, Noor-Ghateh dataset is presented, and finally, a comparison is provided between Noor-Ghateh and the other datasets.
\subsection{Existing Arabic Datasets}
To familiarize ourselves with the features of existing datasets and to compare them with the dataset discussed in this paper, we will introduce and review each one.
\par
\textbf{The Prague Arabic Dependency Treebank \cite{hajic2004prague}} contains multi-level text annotations, including morphological and analytical language representations. It is designed for general use in natural language processing.
This dataset resolves lemmas with the same text representation by adding a number to indicate their meaning.
\par
\textbf{CLARA \cite{zemanek2001clara} corpus} consists of 100,000 words marked morphologically, including strings with defined morphological boundaries. It also includes another annotated corpus. Approximately 15,000 words were analyzed in the CLARA corpus.
\par
\textbf{Penn Arabic Treebank \cite{maamouri2004penn}} contains over half a million (542,543) Arabic words collected from Agence France-Presse, Al-Hayat, and Al-Nahar newspapers. It includes part-of-speech tagging and parsing.
\par
\textbf{Quranic Corpus \cite{zeroual2016new}} contains Quranic Arabic text, with each phrase annotated with morpho-syntactical information.
This corpus includes stems, stem patterns, lemmas, lemma patterns, and roots. It utilizes a semi-automated approach, incorporating the 'AlKhalil Morpho Sys' and manual intervention for accuracy.
\par
\textbf{The Quranic Arabic Corpus \cite{dukes2010morphological}} is an online annotation resource with multiple layers of annotation, including syntactic analysis, dependency grammar, morphological segmentation, POS tagging, and a semantic ontology derived using dependency rules.
\par
\textbf{Al-Mus’haf Corpus \cite{imadmus}} is a Quranic corpus that employs a semi-automated approach using 'AlKhalil Morpho Sys' and guided manual correction informed by morphosyntactic knowledge. It employs a hierarchical structure for Arabic tagset classes and includes translations of the verses into English, Spanish, and French.
\par
\textbf{The Haifa Quranic Corpus (HQC) \cite{dror2004morphological}} includes several analyses for many words.
The HQC authors describe the POS labeling units used to annotate the corpus;
however, the annotation scheme is incomplete and has not been fully manually verified.
As a result, multiple analyses may be applied to each word.
\par
\textbf{Quranic Arabic Corpus \cite{Eric2012corpus}} offers a fine-grained POS tag set for every phrase of the Quran within its contextual verse, all of which have been manually validated, though some noun tags are underspecified. This corpus includes Arabic text alongside phonetic transcription, word-for-word translation, and functional references.
\par
\textbf{The Qur'anic Mushaf collection}, which is not currently available online and is only referenced in \cite{zeroual2014clitiques}, has been enriched with morphological information. Its creation involved a semi-automatic process using Hebron and other procedures. The corpus contains 1,770 roots and specific morphological patterns.
\par
\textbf{Fine-Grained Quran Dataset \cite{hegazi2015fine}} is a Quranic text corpus designed for translation, categorization, parsing, and analysis at multiple levels, including chapter-based and word-based divisions.
It features descriptions, definitions, tone roots, and stems. The final dataset is available in Excel sheets and database document formats.
\par
\textbf{Qatar Arabic Language Bank (QALB) \cite{habash2013qalb} } aims to reach a target size of 2 million words. This extensive dataset supports the development of automatic correction tools for Arabic text through a large collection of manually corrected text, including segmentation.
The project focuses on annotating various types of errors related to morphology and syntax, which requires segmenting text into its constituent parts for detailed analysis and correction.
\par
\textbf{Tashkeela corpus \cite{ZERROUKI2017147}} comprises over 75 million fully vocalized words derived from 97 books, spanning both classical and modern Arabic.
It is widely used to develop and evaluate automatic diacritization systems, crucial for applications like text-to-speech conversion, reading comprehension, and semantic analysis.
While primarily focusing on diacritization, which entails adding diacritical marks to Arabic text for pronunciation, the preprocessing steps often require text segmentation to manage morphological and phonetic elements.
For example, it aids in segmenting words into their prefixes, stems, and suffixes to enhance diacritical accuracy and overall diacritization quality.
\par
\textbf{Arabic Gigaword Fifth Edition \cite{parker2011arabic}} offers an extensive archive of newswire text data from diverse Arabic news sources.
While its primary purpose is not segmentation, researchers often utilize it for broader NLP applications, including segmentation tasks.
This dataset serves as a rich resource for developing and evaluating segmentation algorithms, as it can be segmented into words, phrases, or sentences for various NLP applications.
\par
\textbf{Masader \cite{altaher2022masader}} encompasses datasets annotated for a wide range of NLP tasks, including segmentation.
The Masader catalog features over 600 datasets for Arabic NLP and speech processing, with dataset sizes varying from thousands to millions of words, depending on their scope and intended purpose.

\subsection{Introducing Noor-Ghateh DataSet}
Morphological analysis is one of the most common types of linguistic corpus analysis, as it provides a foundation for other kinds of analysis. 
It can also be performed by computers with high accuracy.
You can choose to adjust the morphological tag to include all morphological features or display them separately.
Arabic words consist of morphemes linked by morphological rules to form basic units.
In this subsection, we present the Noor–Ghateh dataset to evaluate segmentation methods, collected from the book Shariat al–Islam, which contains several chapters on Islamic jurisprudence.
The statistical description of this data is presented in Table \ref{tab1}.

\begin{table}[h]
\caption{Statistical description of Noor-Ghateh dataset}\label{tab1}%
\begin{tabular}{@{}llll@{}}
\toprule
Property & Number \\
\midrule
Number of Jurisprudential Chapters & $52$\\
Number of words	& 223690\\
Number of Sentences	& 10160\\
Number of nouns	& 120432\\
Number of Verbs	& 40029\\
Number of Letters &	124724\\
Number of Root	& 264097\\
Number of Prefix &	74242\\
Number of Suffix & 18617\\
\end{tabular}
\end{table}

\subsubsection{Tags of the Corpus}
\begin{figure}[h]%
\centering
\includegraphics[width=0.9\textwidth]{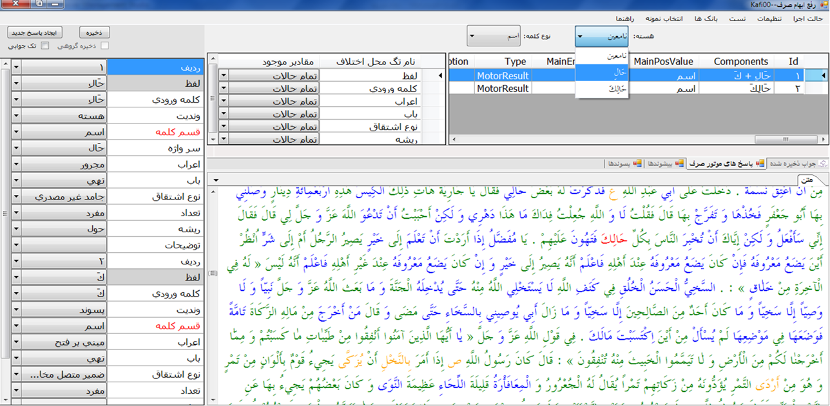}
\caption{A view of the native tool for correcting and revising words}\label{fig1}
\end{figure}

In this dataset, based on the vocabulary analysis, Arabic words are divided and analyzed into three parts: prefix, root, and suffix.
The dataset introduces labels for noun components used in syntactic and morphological analysis, each with a specific meaning and purpose.
Figure \ref{fig1} illustrates the native tool used to correct and correct words in this data set.
There are five distinct groups of labels in this dataset, shown in Table \ref{tab2}.

\begin{table}[h]
\caption{Description of each label}\label{tab2}%
\begin{tabular*}{\textwidth}
{@{\extracolsep\fill}lcccccc}
\toprule%
Property & Number \\
\midrule
Tags &	Meaning/slash values \\
Seq &	order of single words /slash natural numbers\\
Slice &	The content of the single word/slash word \\
Entry &	A morpheme that may contain other morphemes\\
Affix &	Single word type/slash prefix, suffix, nucleus\\
Pos &	Type of word/slash noun, verb, letter\\
\end{tabular*}
\end{table}

\subsubsection{Corpus format}

\begin{figure}[h]%
\centering
\includegraphics[width=0.9\textwidth]{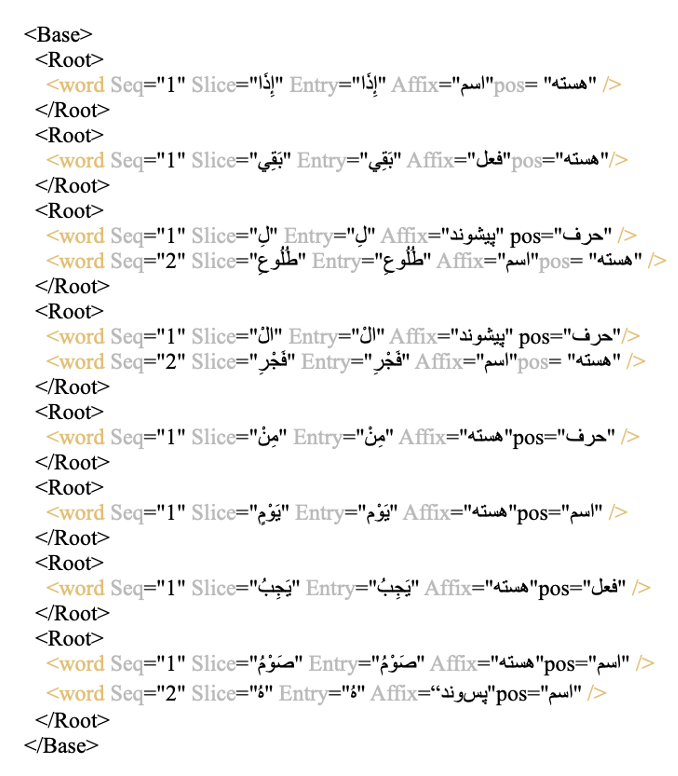}
\caption{Example Analysis encoded in XML format}\label{fig2}
\end{figure}

The corpus was produced in XML format. 
Figure \ref{fig2} shows a sample of the annotated corpus encoded in XML format.
To enhance readability, the Noor–Ghateh dataset was encoded in XML, with words beginning with a $<$Root$>$ tag and ending with a $<$\textbackslash Root$>$ tag.
\par
First, it writes a word sequence of the word, for example, ``\RL{الفجر}'' consists of two parts ``\RL{ال}''+``\RL{فجر}'', so the word sequence for ``\RL{ال}'' is “1” and the word sequence for ``\RL{فجر}'' is “2”, and then the different tags of the word are mentioned according to its type. 

\subsection{Comparison of Arabic Morphological Datasets}
In this subsection, we provide a comparison over the Arabic morphological datasets, including the proposed dataset.

Table \ref{tab3} presents a comparison of different Arabic corpora.

\begin{table}[hp]
\centering
\caption{Comparison between Arabic morphological datasets}
\label{tab3}%
\begin{tabular}{ccccc}
\hline
Dataset & Words & Variety & Annotation for & Topic \\ \hline
Noor–Ghateh Dataset & 223690 & MSA &
\begin{tabular}{@{}c@{}} 
morphological segmentation, \\ POS tagging,\\ lemmatization, \\ diacritization, \\ Root, kind of tagging
\end{tabular}
& Hadith \\ \hline
Prague Arabic \\ Dependency Treebank & 114K & MSA & 
\begin{tabular}{c} 
It is intended for \\  general use  in NLP
\end{tabular}
& News \\ \hline
CLARA & 114K & MSA &
\begin{tabular}{@{}c@{}}
POS tagging 
\end{tabular}
& Varios \\ \hline
Penn Arabic Treebank & 37M & MSA &
\begin{tabular}{@{}c@{}}
tokenization, segmentation, \\POS tagging, lemmatization, \\ diacritization, English gloss \\ and syntactic structure
\end{tabular} 
& News \\ \hline
Quranic Corpus[ zeroual] & more than 1.3M & MSA &
\begin{tabular}{@{}c@{}} 
contains Stem, \\
Stem’s Pattern, \\ 
Lemma, Lemma’s \\
Pattern and Root
\end{tabular}
& Quran \\ \hline
Quranic Arabic Corpus & 17455 Distinct Words & DA+MSA &
\begin{tabular}{@{}c@{}} 
containing syntactic \\ 
evaluation training \\
dependency grammar,\\
morphological segmentation,\\
POS tagging, and a \\
semantic ontology
\end{tabular}
& Quran \\ \hline
Al–Mus’haf Corpus & 77,430 Words & MSA &
\begin{tabular}{@{}c@{}} 
surface words, \\
stems, and lemmas
\end{tabular}
& Quran \\ \hline
Haifa Quranic Corpus & 17455 Distinct Words & MSA &
\begin{tabular}{@{}c@{}} 
Root, Pattern, Words, \\ 
Numbers, Time, \\
Gender Dependent pronouns
\end{tabular}
& Quran \\ \hline
Quranic Arabic Corpus & 77K & MSA &
\begin{tabular}{@{}c@{}} 
Syntactic analysis using \\ dependency grammars and \\ a semantic ontology \\ using morphological segmentation \\ and POS tags
\end{tabular}
& Quran \\ \hline
Qatar Arabic Language Bank & 2M & DA+MSA &
\begin{tabular}{@{}c@{}} 
Tokenization, Segmentation, \\ 
POS Tagging, Lemmatization,\\
Diacritization, English Gloss \\
Syntactic Structure, \\
Stemming, Stem’s Pattern
\end{tabular}
& Varios \\ \hline
Tashkeela & 75M & MSA+CA &
\begin{tabular}{@{}c@{}}
Diacritization, \\
Morphological Analysis \\
Syntax and Grammer
\end{tabular}
&Various  \\ \hline
Arabic Gigaword Fifth Edition & 1.1B & MSA & 
\begin{tabular}{@{}c@{}}
Lexical Analysis, \\ 
Morphological Analysis, \\
Syntactic Parsing, NER
\end{tabular}
& News \\ \hline
Masader & N/A & DA+MSA &
\begin{tabular}{@{}c@{}} 
Dialect Identification, \\
NER, Sentiment Analysis \\
Morphological Analysis
\end{tabular}
& wide range \\ \hline
\end{tabular}
\end{table}

\section{Experiments and Results}
To standardize this dataset as a benchmark, it is necessary to be tested using evaluation methods and analyzed and compared with the other datasets.
This section is dedicated to this purpose.
First, the experiment specifications are provided, and then, the experimented results are presented as well as their analysis.

\subsection{Experiments Specifications}
In this section, first, the experimented segmentation tools are proposed, and then, the other comparison benchmarks are described.
\subsubsection{The Experimented Segmentation Tools}
Farsa, CAMEL, and ALP are the texted segmentation tools, utilized in this paper. This part of the paper is dedicated to describing, more, about them.
\par
\textbf{Farasa.} Farasa \cite{abdelali2016farasa} is a fast and accurate Arabic segmentation tool. Its approach is based on rank SVM and is trained using the Arabic Penn Treebank with linear kernels.
To evaluate Farasa, researchers compared it with two tools: Madamira \cite{pasha2014madamira} and Stanford Arabic \cite{monroe2014word}. Farasa outperforms both in segmentation for information retrieval tasks and is on par with Madamira for machine translation tasks.
\par
\textbf{CAMeL.} The CAMeL system \cite{obeid2020camel} is written in Python and provides a set of tools for managing and modeling Arabic texts.
This system has methods to analyze, generate, and reflect the morphological features of a token, and it supports disambiguation, tagging, tokenization, token recognition, sentiment analysis, and named entity recognition.
It is specially designed for the Arabic language.
\par
\textbf{ALP.} This system \cite{freihat2018single}, has segmentation, POS tagging, and Named Entity Recognition (\,NER)\, responsibilities in a single unmarried process. 
It is designed using OpenNLP.
The annotation process uses three well-known methods: basic annotation to mark the POS, word-phase annotation, and named entity annotation.
The POS tag set contains 58 tags that have been divided into five main classes: noun, adjective, verb, adverb, preposition, and particle. 
The system uses the OpenNLP Maximum Entropy POS tagger with default functionality and a cutoff value of 3.
The corpus used for the assessment was retrieved from the Aljazeera news portal and the Altibbi Science Advisory Network portal, containing 9,990 tokens.
For segmentation and rewriting, they used a maximum entropy tagger, and the research relied on morphological analyzers or dictionaries.
\subsubsection{The Other Tested Benchmarks}
In this part, the NAFIs dataset as well as the Quran dataset are described as the comparison benchmarks, utilized in this paper.
\par
\textbf{NAFIS Dataset.} NAFIS \cite{nafisdataset} (Normalized Arabic Fragments for Inestimable Stemming) is a gold-standard corpus designed for the evaluation of Arabic stemmers, containing 172 words.
It consists of a diverse set of manually annotated sentences that capture the intricacies of Arabic morphology.

NAFIS provides detailed linguistic annotations, making it an invaluable resource for benchmarking and comparing the performance of various stemming algorithms in Arabic natural language processing.
\par
\textbf{Quran Dataset.} The morphological annotation of Quranic Arabic \cite{dukes2010morphological} involves detailed linguistic analysis of the Quran’s text.
This process includes segmenting words into their morphological components, tagging parts of speech, and analyzing syntactic structures using dependency grammar. 
A prominent resource in this field is the Quranic Arabic Corpus, which provides comprehensive annotations for each word in the Quran. 
This corpus includes 7,043 words from the Hamd and Baqarah Surahs.

\subsection{The Experimented Results}

To validate Noor–Ghateh as a benchmark dataset, we evaluated the performance of three methods Farasa, CAMeL, and ALP and measured the corresponding accuracy.

In this evaluation, accuracy is defined as the ratio of the correctly performed separations by the machine to the total separations in the data.
The standard for determining whether segmentation was done correctly is the matching of all segment components with those of the human segmenter.
For example, the machine separation of the phrase ``\RL{سنقتلهم}'' is considered correct when all its components such as
``\RL{س}'',
``\RL{نقتل}'' 
and 
``\RL{هم}'' 
are accurately separated.
\par
Table \ref{tab4} shows the accuracy of word segmentation for three systems, Farasa, CAMEL, and ALP. 
The results were remarkably close.
Performance was assessed based on the aforementioned metrics, revealing closely aligned results across the methods under examination.
Through rigorous evaluation across multiple NLP systems, the Noor–Ghateh dataset demonstrated its versatility and robustness across various NLP tasks.
This comprehensive evaluation not only showcased the dataset compatibility with diverse NLP methodologies but also underscored its reliability and efficacy as a benchmark to evaluate the performance of the system.
Moreover, the meticulous curation process of the Noor–Ghateh dataset (involving expert annotators) guarantees the accuracy and depth of linguistic annotations, providing researchers with a high-quality resource for training and testing NLP algorithms. 
\par
In essence, the Noor–Ghateh dataset emerged as an indispensable asset for Arabic NLP research, offering researchers a perfect blend of authenticity, reliability, and versatility to drive innovation and advancements in the field.
\par
\begin{table}[h]
\caption{The accuracy of word segmentation for the three methods in three datasets}\label{tab4}%
\begin{tabular}{@{}llll@{}}
\toprule
        & Farasa& CAMEL & ALP   \\
\midrule
Nafis   &  0.59 &  0.68 & 0.65  \\
Quran   &  0.76 &  0.80 & 0.81  \\
Sharaye &  0.81 &  0.81 & 0.79  \\
\end{tabular}
\end{table}

The table presents the accuracy of word segmentation for three methods: 
Farasa, CAMEL, and ALP across three datasets: Nafis, Quran, and Sharaye.
The results indicate that CAMEL consistently outperforms the other methods, achieving the highest accuracy in the Nafis dataset (0.68) and tying for the highest accuracy in the Sharaye dataset (0.81).
ALP also performs well, particularly in the Quran dataset where it achieves the highest accuracy (0.81).
Farasa, while generally performing well, demonstrates the lowest accuracy in the Nafis dataset (0.59).
\par
These results highlight the varying effectiveness of each method depending on the dataset, with CAMEL and ALP showing superior performance in most cases.

\section{Conclusion and future work}
This section concludes the paper with final remarks as well as future work suggestions.
\subsection{Concluding remarks}
In this paper, we present the syntactic and morphological data collection of the book, Shariat al–Islam, which is a historical text in the field of hadith.
Human annotations of the text have been collected with the help of experts.
\par
This dataset, characterized by its meticulous curation and fidelity to authentic Arabic language usage, offers an ideal testbed for assessing the capabilities and limitations of various lexical segmentation methodologies.
We will make a sample of this dataset publicly available to support researchers interested in Arabic natural language processing.
\subsection{Future Suggestions}
In subsequent phases of our research, we aim to introduce a Seq-to-Seq method tailored specifically to Arabic lexical segmentation.
This forthcoming method will leverage advanced deep learning architectures and sequence-to-sequence modeling techniques to effectively address the challenges inherent in Arabic language processing, particularly in lexical segmentation tasks.
By harnessing the expressive power of neural networks and incorporating attention mechanisms, our proposed approach aims to enhance the accuracy and robustness of Arabic lexical segmentation, thereby contributing to the advancement of natural language processing methodologies tailored to the Arabic language.
\par
Furthermore, we envision utilizing the presented dataset as a comprehensive benchmark to evaluate the efficacy and performance of the to-be-proposed Seq-to-Seq method, alongside benchmarking against state-of-the-art techniques.
Through rigorous comparative analysis, we seek to provide empirical evidence of the effectiveness and competitiveness of our proposed Seq-to-Seq approach in the realm of Arabic lexical segmentation.

\bibliographystyle{unsrt}  
\bibliography{references}  

\end{document}